\documentclass{article}
\usepackage{spconf,amsmath,graphicx}

\makeatletter
\renewcommand*\env@matrix[1][c]{\hskip -\arraycolsep
  \let\@ifnextchar\new@ifnextchar
  \array{*\c@MaxMatrixCols #1}}
\makeatother

\usepackage{booktabs}
\usepackage{multirow}
\usepackage{siunitx}
\usepackage{amssymb}

\usepackage{float}

\usepackage{amsmath}

\usepackage{todonotes}
\setuptodonotes{color=blue!30}
\title{Hadamard Layer to Improve Semantic Segmentation}
\name{Angello Hoyos, Mariano Rivera} 
\address{Centro de Investigacion en Matematicas, A.C. \\
Guanajuato, Gto., 36023 Mexico \\
\{angello.hoyos, mrivera\}@cimat.mx}

\begin{document}
%
\maketitle
\begin{abstract}

The Hadamard Layer, a simple and computationally efficient way to improve results in semantic segmentation tasks, is presented. This layer has no free parameters that require to be trained. Therefore it does not increase the number of model parameters, and the extra computational cost is marginal. Experimental results show that the new Hadamard layer substantially improves the performance of the investigated models (variants of the Pix2Pix model). The performance's improvement can be explained by the Hadamard layer forcing the network to produce an internal encoding of the classes so that all bins are active. Therefore, the network computation is more distributed. In a sort that the Hadamard layer requires that to change the predicted class, it is necessary to modify $2^{k-1}$ bins, assuming $k$ bins in the encoding. A specific loss function allows a stable and fast training convergence. 

\end{abstract}

\section{Introduction}
\label{sec:intro}

Many problems in computer vision can be defined as image-to-image translation with an input image into a corresponding output image. A clear example is semantic image segmentation, where each pixel in an input image is assigned to a class label; in a certain way, we predict an output image that is less complex than the input, even if the input image is a medical image or a street image seen by an autonomous car. Recently, this task has been solved with different strategies using convolutional neural networks (CNN), i.e., \cite{ulku2022survey, reyes2021w}, even though the main trouble of predicting pixels from pixels does not change.
 
An alternative to solved image segmentation using CNN is Conditional Generative Adversarial Networks (cGAN), an image-to-image translation network. Pix2Pix, proposed by Isola et al. \cite{Pix2Pix}, was the first demonstration of cGANs successfully generating nearly discrete segmentation labels rather than realistic images.

In this work, we present the Hadamard layer. This layer enforces each class label to be encoded to have an equal number of $+1$s and $-1$s, unlike the classic one-hot encoding. With that, we increase the Hamming distance between the class labels and obtain better segmentation results. We think this happens because to attack one-hot encoding requires changing the response in a bin to flip the assigned class, which explains the susceptibility to adversary attacks by neural networks in classification tasks. On the other hand, using the Hadamard encoding requires modifying half of the bins to change a label. In this way, the network requires a more distributed activation to produce a label. 

We present experiments on modified cGANs to evaluate the performance of this new layer. In those experiments, we noted a performance increment without sacrificing training time.

\begin{figure}[htb]

\begin{minipage}[b]{1.0\linewidth}
  \centering
  \centerline{\includegraphics[width=8.5cm]{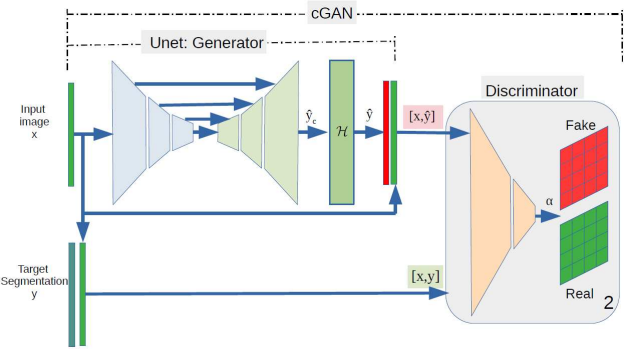}}
\end{minipage}
\caption{Shows a Hadamard Layer embedded in the generator of a Pix2Pix architecture.}
\label{fig:pix2pix_hadamard_img}
\end{figure}

\section{Related Work}
\label{sec:relatedwork}  

\subsection{Semantic Segmentation}
\label{ssec:subhead}

Several convolutional NN architectures for semantic segmentation have been reported in recent years. Among them, popular architectures include UNet \cite{unet}, which was initially developed for BioMedical Image Segmentation. Since its publication in 2015, variants of the UNet have emerged for such purpose. Among them, Pix2Pix-based models are computationally efficient methods that produce good-quality segmentation. Pix2Pix consists of a stable Generative Adversarial Networks (GANs) strategy for training a UNet. The advantage of UNet resides in its simplicity: it consists of a single UNet (hourglass stage). We chose the  Pix2Pix model to introduce our Hadamard Layer in this work. Thus, we evaluate the performance impact of the proposed Hadamard Layer in different implementations, such as the classic UNet, the ResUNet, the VGG-UNet, and the UNet3+ variants. We briefly describe such variants below.

A UNet comprises two stages: a contraction path (encoder) and an expansion one (decoder). The encoder captures the image's context using a stack of convolutional and max pooling layers. The decoder transform combines and expands the extracted features to construct the desired output. ResUNet \cite{resunet} takes the performance gain of Residual networks and uses it with the UNet. This architecture was developed by Zhengxin Zhang et al. and initially used for road extraction from high-resolution aerial images in remote sensing image analysis. Later, it was adopted by researchers for multiple other applications, such as brain tumor segmentation, human image segmentation, and many more. Although this is a capable network, it has a slightly large number of parameters. VGG-UNet \cite{vgg_unet} is another variant of UNet model that combines a pretrained backbone as the encoder. In this proposal, the decoder consists of five up-sampling blocks and uses a symmetric expanding path to enable precise localization segmentation to detect corrosions from steel bridges and cracks from rubber bearings under limited conditions. More recently, UNet3+ \cite{unet3plus} was reported as the best variant for image segmentation: full-scale skip connections incorporate low-level details with high-level semantics from feature maps in different scales and full-scale deep supervision that helps to learn hierarchical representations from the full-scale aggregated feature maps. The growing popularity of the UNet3+ is due to its superior performance compared with other reported variants as Attention UNet \cite{oktay2018attention}, PSPNet \cite{zhao2017pyramid}, and DeepLab \cite{chen2018encoder}, which were considered SOTA.

Despite the advances mentioned in this problem, these solutions are not yet perfect, which is why in this work, we propose a change that can be applied to any neural network that brings a notable benefit. To understand it better, we will first introduce related theory.

\begin{figure*}[!ht]
\centering
\includegraphics[width=1\linewidth]{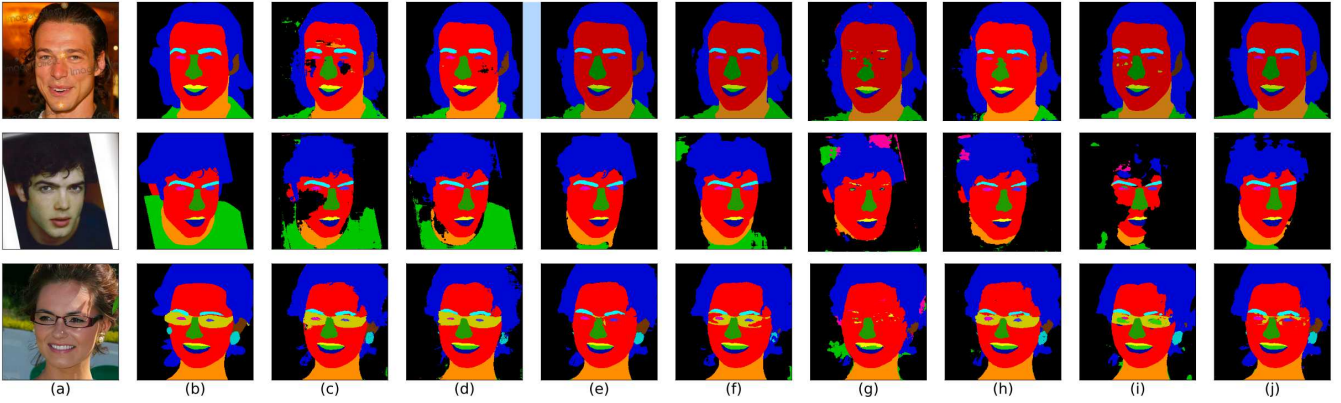}
\caption{Shows some examples of test images from the CelebAMask-HQ dataset: (a) Input image and (b) target segmentation map. The reminder columns depict the results: (c-d) UNet model, (e-f) ResUNet, (g-h) VGG-UNet, and (i-j) UNet3$+$; where columns (c, e, g, i) correspond to models trained with one-hot codification and columns (d, f, h, j) to models trained with Hadamard layer codification.} 
\label{fig:segs}
\end{figure*}

\subsection{Linear Error-Correction Codes}
\label{ssec:linearcodes}

Information transmission is a well-understood area of mathematics where the central idea is to increase the Hamming distance between codewords by adding redundancy to a message to ensure that even if noise flips some random bits, the message is still legible. The solution to this problem is using linear error-correcting codes like Hamming codes \cite{hamming1950error} or Reed-Solomon codes\cite{reed1960polynomial}. In this work, we selected a particular case of Reed-Solomon codes, the Hadamard codes, for our proposal. 

We recognize that Hadamard codes are suitable to be incorporated into neural network architectures, given previous works that show the effectiveness of Hadamard codes in the classification of deep features \cite{deep_hadamard} and as a defense strategy against multiple adversary attacks \cite{hadamard_defense}. Nevertheless, this approach has not been explored in semantic segmentation. 
Hadamard codes are easy to construct; assume we need codes of length $2^k$, or in other words, there are at most $2^k$ possible classes in our segmentation problem; they can be fewer. The Hadamard codes are the rows of the matrix $H_{2^k}$, defined recursively as
\begin{equation}
H_{2^k}=
\begin{bmatrix}[r]
H_{2^{k-1}} & H_{2^{k-1}} \\
H_{2^{k-1}} & - H_{2^{k-1}} 
\end{bmatrix},
\end{equation}
with 
\begin{equation}
H_2=
\begin{bmatrix}[r]
1 & 1 \\
1 & -1 
\end{bmatrix}.
\end{equation}

The above procedure for computing Hadamard matrices is named Silvester's construction and produces symmetric matrices \cite{sylvester}. For example, using this construction for eight classes, we have
\begin{equation}
H_8=
\begin{bmatrix}[r]
1 & \; 1 & 1 & 1 & 1 & 1 & 1 & 1 \\
1 & -1 & 1 & -1 & 1 & -1 & 1 & -1 \\
1 & 1 & -1 & -1 & 1 & 1 & -1 & -1 \\
1 & -1 & -1 & 1 & 1 & -1 & -1 & 1 \\
1 & 1 & 1 & 1 & -1 & -1 & -1 & -1 \\
1 & -1 & 1 & -1 & -1 & 1 & -1 & 1 \\
1 & 1 & -1 & -1 & -1 & -1 & 1 & 1 \\
1 & -1 & -1 & 1 & -1 & 1 & 1 & -1 \\
\end{bmatrix}.
\end{equation}

In the examples above, we can observe that any pair of codes in a $H_n$ Hadamard matrix are at a distance ${2^{k-1}}$; this property allows us to define a more demarcated space between classes, which we consider essential for image segmentation. 

\section{Methodology}
\label{ssec:Methodology}

Figure \ref{fig:pix2pix_hadamard_img} illustrates the use of the proposed \emph{Hadamard Layer} in the UNet model to achieve a segmentation task. Our Hadamard Layer consists of two operations. First, to multiply by $H^\top$ the UNet stage output's  $\hat y_c \in \mathbb{R}^{H \times W \times 2^k}$. Second, to apply a $softmax$ activation to enforce, even more, the output $\hat y$ to be more informative and closer to the one-hot code:
\begin{align}
\label{eq:hadamardlayer}
 \hat y & = \mathcal{H}(\hat y_c) \nonumber \\
        & = \textrm{softmax}( H^\top \hat y_c ).
\end{align}

A valuable property of Hadamard matrices $H$ is  that they are orthogonal:
 $H \,H^\top \; = \; n \,I_n$;
where $n$ is the order, $I_n$ is the $n \times n$ identity matrix and $H^T$ is the transpose of $H$. Since $H$ is orthogonal, $y$ would be closer to a one-hot code if $y_c$ is similar to a hadamard code; \emph{i.e.},  $y_c$ would have all its entries activated. 

Now, we introduce another essential ingredient for having successful training: the loss functions for the cGAN. Then, let $(x, y)$ be the input data for our UNet, $\hat{y}$ is the predicted segmentation, and $\alpha$ the output of the Discriminator (matrix with dimensions $h \times w$) of the full cGAN. Each element $\alpha_{i j}$ can be understood as the probability that the support region of the analyzed pair, $(x,y)$ or $(x,\hat y)$, is real. Then, we use the Discriminator loss given by

\begin{equation}
L_D(\hat{y}) = S(\mathbf{1} | \alpha ) + S(\mathbf{0} | \hat \alpha )
\end{equation}
where $\alpha$ is the discriminator response to the real pair $(x,y)$, and $\hat \alpha$ to the predicted pair $(x,\hat y)$, and $S$ is the cross-entropy loss
\begin{equation}
S(\hat{z}|z) = -\frac{1}{N}\sum_k z_k \, \log \hat{z}_k 
\end{equation}
On the other hand, the loss function for the Generator is
\begin{multline}
\label{eq:gloss}
L_G(\hat{z}) = S(\mathbf{1} | \hat \alpha )   + \Lambda_1 S(\hat{y}|y)  \\
                  + \Lambda_2 MAE(\hat{y}, y) + \Lambda_3 MAE(\hat{y}_c, y_c);
\end{multline}
where the parameters $(\Lambda_1, \Lambda_2, \Lambda_3) = (1000, 100, 250)$ weight the relative contributions of the loss terms, and
\begin{equation}
MAE(\hat{z}, z) = \sum_k |\hat{z}_k - z_k |
\end{equation}
is the mean absolute error. 

Note that the loss \eqref{eq:gloss} includes four terms. The first two terms are the classical one: binary-cross-entropy loss to reward that the generator manages to deceive the discriminator, and a multi-class cross-entropy loss to encourage the predicted segmentation probabilistic map $\hat y$ to be similar to the one-hot encoded ground truth $y$. The third additional loss term reinforces the segmentation by including a $L_1$ penalty for the differences between $\hat y$ and $y$. The last loss term promotes enforcing the input to the Hadamard Layer be similar to the Hadamard code of the corresponding class. We found that the last two loss terms greatly contribute to the stability and convergence ratio of the entire cGAN. In this way, we train our network to compute a \textit{realistic} mask image and return the correct classes. 

\section{Experiments and Results}
\label{sec:experiments}

We evaluate our proposal in the face segmentation task. For this purpose, we use the CelebAMask-HQ (annotation maps). We separated the images into 25,000 training images and 5,000 test images. We investigated the performance improvement when the Hadamard Layers are used in different variants of the Pix2Pix model. Although we have limited our study to Pix2Pix models, it is essential to say that one can introduce the Hadamard Layer in other reported architectures for semantic segmentation. However, such a study is outside the scope of this work. Herein, we investigate the Pix2Pix variants using the standard UNet, the ResUNet, the VGG-UNet, and the UNet3+ networks. Each variant follows a training using one-hot and Hadamard coding for class labels under the same number of steps.

Figure \ref{fig:segs} depicts examples of predicted segmentation for the evaluated models. We compute the segmentation maps $M$ with \begin{equation}
M(r, c) = \underset{k}{\mathrm{argmax}} \, \hat y_k(r, c)
\end{equation}
where $C_k \in {0, 1, 2, ..., K}$ are class index in the matrix 
codes, $(r, c)$ are the pixel at coordinates. We can observe the differences in segmentation for each model in Fig. \ref{fig:segs}. In columns d, f, h, and j, the models that include the Hadamard Layer, we can notice a more detailed segmentation when compared with the respective results in columns c, e, g, and i, the models trained only with one-hot encoding.  

Table 1 resumes the metrics (Pixel Accuracy and the mean Class IoU) of the predicted segmentation for 5,000 test images. In all cases, using a Hadamard Layer gets better results than one-hot codification in these two metrics. Indeed, the models UNet and ResUNet show a more noticeable improvement than VGG-UNet and UNet3+ when using the Hadamard Layer. Layer. 

\begin{table}[ht]
  \begin{tabular}{lSSSS}
    \toprule
    \multirow{2}{*}{Model} &
      \multicolumn{2}{c}{Pixel Acc.} &
      \multicolumn{2}{c}{Class IoU} \\
      & {\small One-Hot} & {\small Hadamard} & {\small One-Hot} & {\small Hadamard} \\
      \midrule
    UNet        & 0.49 & {\bf 0.69} & 0.12 & {\bf 0.23} \\
    ResUNet     & 0.53 & {\bf 0.68} & 0.15 & {\bf 0.28} \\
    VGG-UNet    & 0.61 & {\bf 0.62} & 0.16 & {\bf 0.18} \\
    UNet3+      & 0.64 & {\bf 0.67} & 0.22 & {\bf 0.26} \\
    \bottomrule
  \end{tabular}
  \caption{Accuracy and mean IoU between one-hot and Hadamard codification: CelebAMask-HQ dataset.}
\end{table}

Next, Table 2 presents detailed results for two models, UNet and ResUNet. We can observe some consistency in the IoU values for the class background and face. Meanwhile, there is a quantitative difference favoring the models that use the proposed Hadamard layer for the classes neck, mouth, lips, brow, and hair.

\begin{table}[ht]
  \begin{tabular}{lSSSS}
    \toprule
    \multirow{2}{*}{Class} &
      \multicolumn{2}{c}{UNet} &
      \multicolumn{2}{c}{ResUNet} \\
      & {\small One-Hot} & {\small Hadamard} & {\small One-Hot} & {\small Hadamard} \\
      \midrule
    Background  & 0.76  & {\bf 0.79}  & {\bf 0.81}  & 0.79 \\
    Face        & 0.69  & {\bf 0.76}  & 0.79        & {\bf 0.81} \\
    Neck        & 0.29  & {\bf 0.47}  & 0.40        & {\bf 0.55} \\
    Glasses     & 0.03  & 0.03        & 0.00        & {\bf 0.01} \\
    L-ear       & 0.13  & {\bf 0.14}  & {\bf 0.22}  & 0.19 \\
    R-ear       & 0.10  & {\bf 0.11}  & {\bf 0.20}  & 0.19 \\
    Mouth       & 0.08  & {\bf 0.14}  & 0.09        & {\bf 0.18} \\
    Nose        & 0.10  & {\bf 0.51}  & 0.15        & {\bf 0.61} \\
    U-Lip       & 0.01  & {\bf 0.21}  & 0.03        & {\bf 0.34} \\
    L-Lip       & 0.02  & {\bf 0.30}  & 0.04        & {\bf 0.41} \\
    R-Brow      & 0.01  & {\bf 0.18}  & 0.02        & {\bf 0.27} \\
    L-Brow      & 0.01  & {\bf 0.13}  & 0.03        & {\bf 0.23} \\
    L-Eye       & 0.01  & {\bf 0.11}  & 0.01        & {\bf 0.13} \\
    R-Eye       & 0.00  & {\bf 0.05}  & 0.00        & {\bf 0.06} \\
    Clothes     & 0.00  & {\bf 0.07}  & 0.00        & {\bf 0.08} \\
    Hair        & 0.03  & {\bf 0.47}  & 0.02        & {\bf 0.41} \\
    Earrings    & 0.00  & 0.00  & 0.00  & 0.00 \\
    Necklace    & 0.00  & 0.00  & 0.00  & 0.00 \\
    Hat         & 0.00  & 0.00  & 0.00  & 0.00 \\
    \bottomrule
  \end{tabular}
  \caption{Pixel Accuracy and Class IoU between one-hot and Hadamard class codification for test Images in CelebAMask-HQ dataset.}
\end{table}

The Discriminator of all the evaluated models is of the PatchGAN type. The Discriminator's output is a matrix where each element evaluates the realness of the support region (receptive field). According to Isola et al., for generating images from semantic segmentations in the CitySky database, the appropriate receptive field is $70 \times 70$ pixels \cite{Pix2Pix}. For our task, we choose the receptive field size as that one produces the best results in the model based on the simple UNet network with one-hot encoding. We conducted this experiment on a subset of just 1000 training images and 100 testing images; instead of the complete data set of 25,000/5000 for training/testing images. The receptive field size turned out to be $30 \times 30$. In Table 3, we show the results for different receptive field sizes. Note that models with the Hadamard layer perform better for larger receptive fields. However, we use a patch size equal to $30 \times 30$ for a fair comparison.

\begin{table}[ht]
  \begin{tabular}{lSSSS}
    \toprule
    \multirow{2}{*}{Patch Sz} &
      \multicolumn{2}{c}{Pixel Acc.} &
      \multicolumn{2}{c}{Class IoU} \\
      & {\small One-Hot} & {\small Hadamard} & {\small One-Hot} & {\small Hadamard} \\
      \midrule
    15x15 & 0.51 & 0.66 & 0.12 & 0.21 \\
    {\bf 30x30} & 0.54 & 0.64 & 0.14 & 0.20 \\
    60x60 & 0.51 & 0.66 & 0.12 & 0.19 \\
    120x120 & 0.51 & 0.67 & 0.13 & 0.21 \\
    \bottomrule
  \end{tabular}
  \caption{Pixel Accuracy and Class IoU between one-hot and Hadamard class codification for test Images in CelebAMask-HQ Mini dataset, with different patch sizes in the Discriminator.}
\end{table}

\section{Conclusions and future work}
\label{sec:conclusions}

We proposed the Hadamard Layer as a simple and computationally efficient way to improve results in semantic segmentation tasks. The new layer is constant, so it does not increase the number of model parameters. As test architecture, we use different variants of the Pix2Pix model for the face segmentation task using the CelebAMask-HQ database. The results show that the new Hadamard layer substantially improves the performance of the investigated models. The metrics evaluated are the simple accuracy (number of pixels correctly classified) and the Intersection Over Union (IOU). The best performance of the intervened models can be explained by the Hadamard layer forcing the network to produce an encoding of the classes so that all bins are active. Consequently, the network computation is more distributed. In a sort that the Hadamard layer requires that to change the predicted class, it is necessary to modify $2^{k-1}$ bins, assuming $k$ bins in the encoding.
On the other hand, changing the predicted class in one-hot encoding is enough to modify a single bin (decrease the response of the most significant bin or increase the response of any other.). 
Our future work will extend our evaluation to other databases and different architectures and assess the performance of implicit coding against adversary attacks.

{\bf Acknowledgments}. This work was partly supported by the CONACYT (Mexico); Grant CB 2017-2018-A1-S-43858.

\bibliographystyle{IEEEbib}
\bibliography{strings, refs}

\end{document}